\pdfoutput=1
\PassOptionsToPackage{table,dvipsnames}{xcolor}
\documentclass[11pt]{article}

\usepackage[preprint]{latex/acl}

\usepackage{times}
\usepackage{latexsym}

\usepackage[T1]{fontenc}

\usepackage[utf8]{inputenc}

\usepackage{microtype}

\usepackage{inconsolata}

\usepackage{graphicx}
\usepackage{hyperref}
\usepackage{multirow}
\usepackage{multicol}
\usepackage{adjustbox}
\usepackage{amsmath}
\usepackage{ascii}
\usepackage{subcaption}  
\usepackage{booktabs}
\usepackage{todonotes}
\usepackage{soul} 
\usepackage{marginnote}
\usepackage{url}
\usepackage{pifont}

\usepackage[capitalize]{cleveref}
\Crefname{appendix}{App.}{Apps.}
\usepackage{listings}
\definecolor{TodoColor}{rgb}{1,0.7,0.6}
\definecolor{TodoColor2}{rgb}{0.7,0.7,0.9}
\definecolor{TodoColor3}{rgb}{0.5,0.8,0.5}

\definecolor{coquelicot}{rgb}{1.0, 0.22, 0.0}

\usepackage{tikz}
\definecolor{mygold}{RGB}{255, 215, 0} 
\definecolor{mysilver}{RGB}{192,192,192} 
\definecolor{mydarkgreen}{RGB}{0, 100, 0}
\newcommand{\gold}{\,\raisebox{0.6ex}{\tikz\draw[fill=mygold,draw=mygold] (0,0) circle (0.55ex);}}
\newcommand{\nosplit}{\,\raisebox{0.6ex}{\tikz\draw[line width=1.4pt, draw=blue] (0,0) circle (0.55ex);}}
\newcommand{\vad}{\,\raisebox{0.6ex}{\tikz\draw[fill=mysilver,draw=mysilver] (0,0) circle (0.55ex);}}

\title{KIT's Offline Speech Translation and Instruction Following Submission for IWSLT 2025}

\author{
 \textbf{Sai Koneru$^*$\textsuperscript{\ENQ}}, 
 \textbf{Maike Züfle$^*$\textsuperscript{\ETX}}, 
 \textbf{Thai-Binh Nguyen}, 
 \textbf{Seymanur Akti}, 
 \textbf{Jan Niehues}, \\
 \textbf{Alexander Waibel} \\
 \\
 Karlsruhe Institute of Technology \\
 \\
   \href{mailto:email@domain}{firstname.lastname@kit.edu}
}

\begin{document}
\maketitle
\def\thefootnote{}\footnotetext{* Equal Contribution}\def\thefootnote{\arabic{footnote}}
\def\thefootnote{}\footnotetext{\ENQ Offline, \ETX Instruction-Following}\def\thefootnote{\arabic{footnote}}
\begin{abstract}
The scope of the International Workshop on Spoken Language Translation (IWSLT) has recently broadened beyond traditional Speech Translation (ST) to encompass a wider array of tasks, including Speech Question Answering and Summarization. This shift is partly driven by the growing capabilities of modern systems, particularly with the success of Large Language Models (LLMs). In this paper, we present the Karlsruhe Institute of Technology's submissions for the Offline ST and Instruction Following (IF) tracks, where we leverage LLMs to enhance performance across all tasks.
For the Offline ST track, we propose a pipeline that employs multiple automatic speech recognition systems, whose outputs are fused using an LLM with document-level context. This is followed by a two-step translation process, incorporating additional refinement step to improve translation quality. For the IF track, we develop an end-to-end model that integrates a speech encoder with an LLM to perform a wide range of instruction-following tasks. We complement it with a final document-level refinement stage to further enhance output quality by using contextual information. 

\end{abstract}

\section{Introduction}

This paper provides an overview of the systems submitted by the Karlsruhe Institute of Technology (KIT) to the \href{https://iwslt.org/2025/offline}{Offline Speech Translation} (ST) and the Constraint Long \href{https://iwslt.org/2025/instruction-following}{Instruction-Following} (IF) tasks of IWSLT 2025. For the Offline track, we participate in the unconstrained setting for the \textit{English}$\rightarrow$\textit{German} language pair. For the IF task, we participate in the constrained-long track, aiming to perform Automatic Speech Recognition (ASR), Speech Translation (ST), Spoken Question Answering (SQA), and Speech Summarization (SSUM) across various languages.

A growing research trend in the field is the application of Large Language Models (LLMs) to speech processing tasks \citep[among others]{tangsalmonn, züfle2024contrastivelearningtaskindependentspeechllmpretraining, chu2024qwen2, abouelenin2025phi}, leveraging their strong general knowledge and natural language understanding capabilities. These strengths make LLMs particularly relevant to both the Offline ST and IF tracks. Accordingly, in our submissions, we explore strategies for effectively integrating LLMs into speech processing pipelines.

There are multiple approaches to leveraging LLMs in speech systems. One strategy involves incorporating LLMs as an additional step within a cascaded architecture \citep{koneru-etal-2024-blending}, where they can perform task-specific refinement. This modular approach allows each component to be trained independently, benefiting from specialized data. Alternatively, LLMs can be integrated in an end-to-end fashion \citep{tangsalmonn, züfle2024contrastivelearningtaskindependentspeechllmpretraining, chu2024qwen2, abouelenin2025phi}, allowing for better information flow and potentially improving generalization to unseen tasks.

\begin{figure*}[ht]
    \centering
    \includegraphics[width=0.9\linewidth]{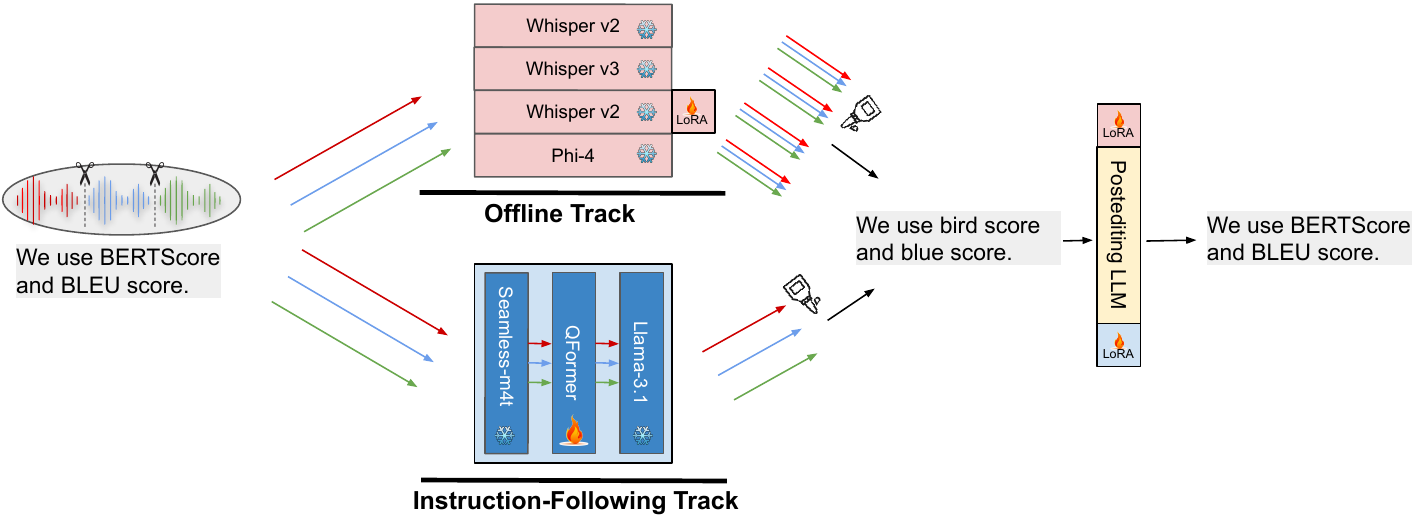}
    \caption{For the Instruction-Following track, we train an end-to-end SpeechLLM, while the Offline track relies on an ensemble of existing models. To enhance the outputs from both tracks, we apply a post-editing model that provides two main benefits: correcting scientific terminology and recovering context that may be lost due to the segmentation of long audio sequences.}
    \label{fig:postedit}
\end{figure*}

Although both the Offline and IF tasks fall under the umbrella of speech processing, they differ significantly in nature. In the offline setting, speed and adaptability to unseen tasks are not primary concerns. In contrast, the IF task demands flexibility and generalization, as the system must handle a variety of instructions. This has an impact on the architectures we choose for the different tracks.

For the Offline track, we utilize LLMs specialized on a specific task as refinement modules within a cascaded architecture. This is common practice; all systems submitted to IWSLT 2024 for this track employed a cascaded architecture \citep{ahmad-etal-2024-findings}, underlining its practical advantages in training the system due to availability in data, e.g for low-resource languages \citep{liu-etal-2023-kits}, and simplicity by decomposing into smaller tasks. 

For the IF track, training a dedicated cascaded system for each task is not an efficient solution, moreover, the goal of this track is to build a model that can follow different instructions. Consequently, we adopt an end-to-end approach using a Speech Large Language Model (SpeechLLM). Nevertheless, for tasks such as ASR, ST, and SSUM, we also include an additional refinement step to enhance fluency and contextual consistency in the output.

An overview of both systems exploiting LLMs internally or via post-editing for refinement, can be found in \cref{fig:postedit}. We describe the details of each system in the following sections. First, we present the Offline ST track system in \cref{sec:offline}. Then, we discuss the IF track system in \cref{sec:IF}.

\section{Offline Track}
\label{sec:offline}
The goal of the Offline ST track is to generate high-quality translations across diverse domains without latency constraints. Recent work has highlighted the potential of LLMs for this task \citep{ahmad-etal-2024-findings, koneru-etal-2024-blending}. Building on these insights, we integrate LLMs at multiple stages of our speech translation pipeline. Below, we present a high-level overview, with each component detailed in the following sections.

We begin with long-form audio inputs, which may span several minutes to hours. Due to memory limitations and the lack of training data for such durations, our ASR and MT systems cannot handle these directly. Thus, we first segment the audio into manageable chunks using a Voice Activity Detection (VAD)-based method, which is effective even in noisy conditions.

The segmented audio is then transcribed into English using ASR. Rather than relying on a single model, we adopt a fusion strategy, combining outputs from multiple ASR systems—including both pre-trained models and a fine-tuned variant. This approach, akin to model ensembling, leverages the complementary strengths of different systems to reduce errors.

We fuse the ASR outputs using an LLM, which processes the combined hypotheses at the document level. This allows for the incorporation of broader context, resulting in more coherent and accurate transcriptions.

The English text is then segmented into sentences using the \texttt{nltk} tokenizer and translated into German. For this, we fine-tune a translation LLM on high-quality parallel data. To ensure quality, we use a quality estimation model to filter out noisy sentence pairs, keeping only high-confidence examples.

Finally, both the source transcript and the machine-translated output are passed to an Automatic Post-Editing (APE) model. This model refines the translations, producing polished final outputs. 

\subsection{Segmentation}

The segmentation module breaks long-form audio into manageable segments for the ASR pipeline. We explored two strategies: fixed-window chunking and content-aware segmentation.

Fixed-window chunking applies a uniform sliding window and relies on transcript overlap to stitch adjacent chunks. While effective on clean audio, it often fails in noisy settings like the ITV or EPTV datasets, leading to fragmented or duplicated text.

Content-aware segmentation uses audio cues to find natural cut points. Basic methods rely on VADs like Silero \citep{SileroModels} or py-webrtcvad \citep{wiseman2019wiseman}, which work well in clean conditions but struggle with noise. Instead, we use an end-to-end speaker segmentation model from \citet{bredin21_interspeech}, trained for noisy scenarios and capable of tracking up to three speakers. While methods like SHAS \citep{tsiamas22_interspeech} use wav2vec embeddings, they underperform in the presence of background noise.

Even with smarter cut-point detection, uncontrolled segment lengths can hurt ASR performance. Inspired by WhisperX \citep{bain2022whisperx}, we enforce length constraints by post-processing VAD segments: overly long segments are split at their lowest-confidence point, while overly short ones are merged with neighbors (even across non-speech gaps) until they reach the desired duration \textit{Chunk Size}.

\begin{table}[!ht]
\centering
\begin{tabular}{@{}c|cccc@{}}
\toprule
\multicolumn{1}{l|}{Chunk Size} & \textit{Peloton} & \textit{EPTV}  & \textit{ITV}   & \textit{ACL}   \\ \midrule
5                               & 13.62            & 15.79          & 21.49          & 14.38          \\
10                              & 12.61            & 14.63          & 18.8           & 12.03          \\
15                              & 12.23            & 14.08          & 17.71          & \textbf{11.43} \\
20                              & 12.27            & \textbf{13.98} & 17.29          & 11.71          \\
25                              & \textbf{11.98}   & \textbf{13.98} & \textbf{16.62} & 11.49          \\ \bottomrule
\end{tabular}
\caption{Impact of chunk size during segmentation for ASR. We report the WER scores using Whisper-v3 with different chunk sizes. Best scores for each test set are highlighted in \textbf{bold}.}
\label{tab: seg_asr}
\end{table}

To determine the optimal \textit{chunk size}, we perform a grid search using test sets from various domains, with results shown in Table \ref{tab: seg_asr}. We use the \href{https://huggingface.co/openai/whisper-large-v3}{Whisper v3 model}\footnote{\texttt{openai/whisper-large-v3}} \citep{radford2023robust} and evaluate it on the Peloton, EPTV, and ITV subsets from the IWSLT 2024 development sets \citep{ahmad-etal-2024-findings}, as well as the ACL 60/60 test set \citep{salesky-etal-2023-evaluating}. A chunk size of $25$ consistently yields the best performance. We hypothesize that this is due to the larger chunk size offering more contextual information, aligning with prior work on the benefits of long-form decoding in noisy conditions \citep{koneru-etal-2024-blending,yan-etal-2024-cmus}.

\subsection{Automatic Speech Recognition}

After segmenting the audio into smaller chunks, we send them to the ASR system for transcription. Since we participated in the language direction \textit{English}$\rightarrow$German, the audio needs to be transcribed in English, a high-resource language. Many publicly available pre-trained models excel at English transcription, and we first evaluated several of them individually. Specifically, we considered the Whisper variants v2\footnote{openai/whisper-large-v2} and v3\footnote{openai/whisper-large-v3}\citep{radford2023robust}, as well as the recently developed multimodal LLM Phi-4\footnote{microsoft/Phi-4-multimodal-instruct}\citep{abouelenin2025phi}.

To build a robust model for noisy scenarios, such as those found in TV series, we further fine-tuned Whisper Large v2 on the Bazinga dataset \citep{lerner-etal-2022-bazinga}. The Word-Error-Rate (WER) for these models on ITV and ACL 60/60 are reported in \cref{tab:all_asr}. 

\begin{table}[!ht]
\centering
\begin{tabular}{@{}c|cc@{}}
\toprule
Model              & \textit{ITV} & \textit{ACL} \\ \midrule
Whisper v2         & 17.04        & 11.55        \\
Whisper v2 + Bazinga & 16.87        & 11.23        \\
Whisper v3         & \textbf{16.62}        & 11.49        \\
Phi-4              & 20.64        & \textbf{9.71}         \\
LLM-Fuse           & 17.03   & 10.77        \\ \bottomrule
\end{tabular}
\caption{WER scores of ASR models on the ITV and ACL test sets. LLM-Fuse indicates the post-edited output of all ASR systems at document-level. Best scores for each test set are highlighted in \textbf{bold}.}
\label{tab:all_asr}
\end{table}

As shown in \cref{tab:all_asr}, there is no clear winner across the two test sets. Our manual analysis further reveals that different models tend to make different types of errors, suggesting that combining these systems could be a promising strategy.

\subsubsection{Fusing with LLM}
\label{subsec:SHORT_asrpostedit}
To fuse the ASR outputs, token-level ensembling is a viable approach—provided the vocabularies of the systems are compatible. However, the vocabulary used by Phi-4 differs from that of the Whisper variants, limiting the effectiveness of this method. Alternative techniques such as re-ranking offer some promise but are unable to leverage document-level context.

To overcome these limitations, we employ an LLM to generate the final transcript based on the outputs from individual ASR systems. Thanks to their ability to process long contexts, LLMs enable us to concatenate hypotheses from multiple chunks and refine them collectively.

However, an off-the-shelf LLM may not perform optimally for this specific task. To improve, we propose fine-tuning the model using a dataset generated through data augmentation. For this purpose, we use monolingual English text from the Europarl v7 and v10 datasets \citep{koehn-2005-europarl}, NewsCommentary v16, OpenSubtitles \citep{lison-tiedemann-2016-opensubtitles2016}, and the NUTSHELL dataset\footnote{Our submission is unconstrained by using this dataset.} \citep{zufle2025nutshell}. With the exception of NewsCommentary, the other datasets contain document-level structure—episodes in the case of OpenSubtitles and abstracts in the case of NUTSHELL.

We then employ the Text-to-Speech model VITS \citep{kim2021conditional} to synthesize audio from the selected texts. This generated audio is subsequently transcribed using Phi-4 and the Whisper variants. As a result, we obtain ASR hypotheses for the synthesized speech along with their corresponding ground-truth references.

Next, we convert this data into a prompt format, as described in Appendix \cref{sec:fuseasrprompt}. We fine-tune the LLM Llama 3 8B\footnote{meta-llama/Llama-3-8B}\citep{grattafiori2024llama3herdmodels} using LoRA \citep{hulora}, training it to predict the reference transcription given the hypotheses produced by the different ASR systems. We illustrate this in \cref{fig:postedit}. We also report the ASR performance of the LLM fusion approach in \cref{tab:all_asr} and observe that it does not outperform the individual systems. However, as we demonstrate in the following sections, this fusion proves to be highly beneficial when computing the final ST scores.

\subsection{Speech Translation}

The next step in the pipeline, after performing ASR, is to translate the transcriptions into German. Since the transcriptions are produced at the chunk level, they often contain multiple sentences, some of which may be incomplete. To address this, we first concatenate all the text from a given talk and then segment it into sentences using the NLTK tokenizer. This ensures that only complete sentences are passed to the MT system, aligning with the way such systems are typically trained.

\subsubsection{Gold vs ASR Transcripts}
\label{sec:gold_asr}

\begin{table}[!ht]
\resizebox{1\columnwidth}{!}{
\centering
\begin{tabular}{@{}cccc@{}}
\toprule
\multicolumn{1}{c|}{Model}     & \textit{Chrf2 ($\uparrow$)} & \textit{MetricX ($\downarrow$)} & COMET ($\uparrow$) \\ \midrule
\multicolumn{4}{c}{Gold Transcript \gold}                                                                                 \\ \midrule
\multicolumn{1}{c|}{Tower 7B}  & 68.7                        & 2.02                            & 83.31              \\
\multicolumn{1}{c|}{ GemmaX2 9B} & 70.5                        & 2.08                            & 83.62              \\ \midrule
\multicolumn{4}{c}{Whisper v3 ASR (Chunk size=25) \vad}                                                                  \\ \midrule
\multicolumn{1}{c|}{Tower 7B}  & 66.1                        & 2.46                            & 81.01              \\
\multicolumn{1}{c|}{ GemmaX2 9B} & 66.4                        & 2.65                            & 80.74              \\ \midrule
\multicolumn{4}{c}{Phi-4 ASR \vad}                                                                                       \\ \midrule
\multicolumn{1}{c|}{Tower 7B}  & 64.9                        & 2.73                            & 79.25              \\
\multicolumn{1}{c|}{ GemmaX2 9B} & 65.4                        & 2.9                             & 79.12              \\ \bottomrule
\end{tabular}
}
\caption{Translation quality comparison between Gold and ASR transcripts on the ACL 60/60 test set. Note that higher is better for chrf2 and COMET scores and lower for MetricX scores.}
\label{tab:st_gold}
    \vspace{-0.25cm}
\end{table}

Recently, several translation-focused LLMs have been introduced, demonstrating strong performance on high-quality input \citep{xu2024x, alvestower}. However, their effectiveness on noisy input—such as ASR-generated transcripts—remains uncertain. To assess this, we first evaluate the out-of-the-box translation quality of two leading models: Tower\footnote{Unbabel/TowerInstruct-7B-v0.2} \citep{xu2024x} and GemmaX2\footnote{ModelSpace/GemmaX2-28-2B-v0.1} \citep{cui2025multilingual}. We use the COMET\footnote{Unbabel/wmt22-comet-da} \citep{rei-etal-2022-comet}, MetricX\footnote{google/metricx-24-hybrid-xl-v2p6} \citep{juraska-etal-2024-metricx}, and ChrF2 \citep{popovic-2015-chrf} metrics, with results reported in \cref{tab:st_gold} for the ACL 60/60 test set.

GemmaX2 outperforms Tower on gold transcripts in terms of COMET scores, but its performance drops significantly on ASR-generated input. Interestingly, translation quality is lower when using transcripts from the Phi-4 ASR model, despite it having the lowest WER in \cref{tab:all_asr}. We hypothesize that this is due to inconsistencies in punctuation and casing, which are not captured by WER but can impact translation quality. This highlights that lower WER does not always correlate with better translations. As a result, we choose Tower 7B as our base model for subsequent enhancements, given its superior robustness to noisy input.

\begin{table*}[!ht]
\resizebox{2\columnwidth}{!}{
\begin{tabular}{@{}cccccc@{}}
\toprule
\multicolumn{1}{c|}{\multirow{2}{*}{Model}}             & \multicolumn{2}{c}{ITV}                                       & \multicolumn{3}{c}{ACL}                                                            \\ \cmidrule(l){2-6} 
\multicolumn{1}{c|}{}                                   & \textit{Chrf2 ($\uparrow$)} & \textit{MetricX ($\downarrow$)} & \textit{Chrf2 ($\uparrow$)} & \textit{MetricX ($\downarrow$)} & COMET ($\uparrow$) \\ \midrule
\multicolumn{6}{c}{Whisper v3 ASR (Chunk size=25) \vad}                                                                                                                                                           \\ \midrule
\multicolumn{1}{c|}{Tower 7B}                           & 41.4                        & 4.25                            & 66.1                        & 2.46                            & 81.01              \\
\multicolumn{1}{c|}{Tower 7B Finetuned}                 & 41.5                        & 4.19                            & 67.7                        & 2.27                            & 82.05              \\ \midrule
\multicolumn{6}{c}{LLM-Fuse \nosplit}                                                                                                                                                                                 \\ \midrule
\multicolumn{1}{c|}{Tower 7B Finetuned}                 & 41.7                        & 4.12                            & 68                          & 2.01                            & 83.07              \\
\multicolumn{1}{c|}{Tower 7B Finetuned + Tower 13B APE} & 42.1                           & 4.03                               & 69.6                        & 1.84                            & 83.31              \\ \bottomrule
\end{tabular}
}
\caption{Analysis of translation quality of our ST system with different enhancements on the ITV and ACL test sets. Note that higher is better for chrf2 and COMET scores and lower for MetricX scores. Best scores for each metric per test set are highlighted in \textbf{bold}.}
\label{tab:st_all}
    \vspace{-0.25cm}
\end{table*}

\subsubsection{Quality-Filtered Finetuning for MT}

Tower 7B is a multilingual model and we only focus on English $\rightarrow$ German in our submisison. Therefore, we adapt it to this specific language pair. While plently of data is available for fine-tuning, these also include low quality translation pairs. 

Recent studies have demonstrated the importance of high-quality data during fine-tuning \citep{finkelstein-etal-2024-introducing, ramos-etal-2024-aligning, xu2024contrastive}. To this end, we leverage the Europarl v7 and v10 datasets \citep{koehn-2005-europarl}, NewsCommentary v16, and OpenSubtitles \citep{lison-tiedemann-2016-opensubtitles2016} to extract high-quality translation pairs. We employ the XCOMET\footnote{Unbabel/XCOMET-XL} quality estimation model \citep{guerreiro2024xcomet} to rank the translation pairs and select the top $500k$ based on quality scores. Tower 7B is then fine-tuned on this curated dataset using LoRA adapters \citep{hulora}, adapting it for generating German translations.

\subsubsection{Automatic Post-Editing Translations}\label{subsec:SHORT_postedit}
As a final step, we aim to correct translation errors through APE \citep{koneru-etal-2024-contextual}. To achieve this, we fine-tune Tower 13B\footnote{Unbabel/TowerInstruct-13B-v0.1} on a synthetically generated APE dataset. Using our previously fine-tuned model, we generate 100k \textit{(source, hypothesis, reference)} triplets by sampling a subset from the top 500k high-quality sentence pairs. Then, we transform into the prompt format as shown in \cref{sec:fuseasrprompt}. We choose the larger 13B model for this task, as we expect it to be adaptable to correct the output with limited fine-tuning. To train within resource constraints, we follow the same approach as before and fine-tune using LoRA adapters.

We present an overview of the ST scores in \cref{tab:st_all} for the ITV and ACL 60/60 test sets. The results show that fusing system hypotheses using an LLM leads to improved ST performance on both test sets (from $4.19 \rightarrow 4.12$ for ITV and $2.27 \rightarrow 2.01$ for ACL in MetricX). Additionally, applying Automatic Post-Editing (APE) further enhances translation quality. As a result, our final pipeline integrates multiple ASR systems fused via an LLM, followed by initial translation generation and post-editing to ensure high-quality output.

\subsection{Future Directions and Potential Improvements}

There are several potential avenues for improving our approach in future iterations of the shared task. First, while we did not explore it in this work, it is unclear how well SHAS segmentation performs when trained on noisy data. Semantic segmentation of noisy inputs could yield performance gains. Second, incorporating LLM specific to the target language (e.g. German LLM) for APE at the document level could offer promising improvements. Lastly, we experimented with Quality-Aware Decoding \citep{koneru2025quality}, which showed benefits primarily when the quality of the ASR output was high. Future research could focus on adapting the quality estimation component to perform robustly under noisy or imperfect segmentation conditions.

\section{Instruction Following Long Track}\label{sec:IF}
The Instruction-Following (IF) Speech Processing track in the scientific domain aims to benchmark foundation models that can follow natural language instructions—an ability well-established in text-based LLMs but still emerging in speech-based counterparts. The track covers four tasks: Automatic Speech Recognition (ASR), Speech Translation (ST), Spoken Question Answering (SQA), and Spoken Summarization (SSUM). ASR is evaluated on English, ST on English $\to$ German, Chinese, and Italian (en$\to$\{de, it, zh\}), and SQA/SSUM across all four directions (en$\to$\{en, de, it, zh\}).

We participate in the Constrained Long track, which focuses on long-form speech inputs (5–10 minutes). This track enforces limitations on both model selection and training data. Specifically, only SeamlessM4T-Large\footnote{\label{foot:seamless}facebook/seamless-m4t-v2-large} \citep{seamlessm4t2023} and LLaMA-3.1-8B-Instruct \footnote{\label{foot:llama}meta-llama/Llama-3.1-8B-Instruct} \citep{grattafiori2024llama3herdmodels} are permitted as base models. 

Our approach employs an end-to-end speech model trained under these constraints,  enhanced with a post-editing stage for improved output quality similar to the Offline track.

\subsection{Data}\label{subsec:if_data}
\paragraph{Data in the Constrained Setting}
For ASR and ST, the provided datasets include EuroParl-ST \citep{jairsan2020a} and CoVoST 2 \citep{wang2020covost}. For the SQA task, the only resource available is the extractive Spoken-SQuAD \citep{lee2018spoken}. For SSUM, NUTSHELL \citep{züfle2025nutshelldatasetabstractgeneration}, an abstract generation dataset for scientific talks, is provided. As development data, the ACL 60/60 benchmark \citep{salesky-etal-2023-evaluating} is made available. Notably, the only in-domain datasets, i.e., those based on scientific talks, are NUTSHELL and ACL 60/60. Moreover, no multilingual data is provided for SQA and SSUM.

\paragraph{Data Augmentation}
To address the limitations of the constrained setting, we apply task-specific data augmentation strategies\footnote{Augmented Dataset available at HuggingFace: \href{https://huggingface.co/datasets/maikezu/data-kit-sub-iwslt2025-if-long-constraint}{maikezu/data-kit-sub-iwslt2025-if-long-constraint}}:

\textbf{ASR:} To introduce domain-specific data, we augment the ASR training data using scientific abstracts from  NUTSHELL \citep{züfle2025nutshelldatasetabstractgeneration}. The abstracts are split into sentences with \texttt{nltk} and then converted to synthetic speech using SeamlessM4T-Large. 

\textbf{ST:} We do not augment the ST training data, but construct an artificial en-it test set for the ACL 60/60 dataset, which lacks Italian. We translate the English ACL 60/60 transcripts into Italian using both SeamlessM4T-Large and LLaMA-3.1-8B-Instruct, and evaluate translation quality using COMETKiwi \citep{rei-etal-2022-cometkiwi}. SeamlessM4T-Large achieves a slightly higher score (82.55 vs. 81.07), and is therefore used to generate the final test set translations. The translation prompts for LLaMA-3.1-8B-Instruct are detailed in \cref{app:data_augm_prompts_st}.

\textbf{SQA:} For SQA, we aim to: (1) support all language pairs, (2) adapt to the scientific domain, and (3) include abstractive QA, as required by the track. Therefore, we transcribe NUTSHELL dev talks using SeamlessM4T (audio split into 15-second chunks at silence regions). We then use LLaMA-3.1-8B-Instruct to generate two answerable and one unanswerable QA pair per segment for all language pairs. We  balance the dataset by ensuring that unanswerable questions comprise 5\% of the final set.  Additionally, we generate a 250-sample test set from a subset of the NUTSHELL test data. Prompt templates are included in \cref{app:data_augm_prompts}

\textbf{SSUM:} To enable multilingual evaluation of speech summarization, we translate the full NUTSHELL dataset (en$\to$\{de, it, zh\}) using LLaMA-3.1-8B-Instruct. Prompt details are provided in \cref{app:data_augm_prompts_ssum}. As with SQA, we also generate a 250-sample multilingual test set.

\subsection{Model}\label{subsec:if_model}

In the constrained setting of the track, only the speech foundation model  SeamlessM4T-Large\footref{foot:seamless} \citep{seamlessm4t2023}  and LLaMA-3.1-8B-Instruct\footref{foot:llama} \citep{grattafiori2024llama3herdmodels} are permitted.

\paragraph{Architecture}
To integrate the speech encoder and LLM in an end-to-end architecture, we use  Q-Former \citep{Li2023BLIP2BL, tang2024salmonn} as a projector. Specifically, we use a four transformer layers and four learnable query tokens to bridge the modality gap between the features from SeamlessM4T and LLaMA. During training, only the projector is trained and the speech encoder and LLM remain frozen.

\paragraph{Training}
\begin{table*}[t]
    \centering
    \begin{tabular}{l|c|ccc|c|c}

    \toprule
     \multirow{4}{*}{\textbf{Model}}&  \textbf{ASR}\gold & \multicolumn{3}{c|}{\textbf{ST}\gold} & \textbf{SQA}\nosplit & \textbf{SSUM}\nosplit \\

        &  ACL 60/60 & \multicolumn{3}{c|}{ACL 60/60} &  Sp.-SQuAD &  NUTSHELL\\
         &  WER & \multicolumn{3}{c|}{COMET} &  BERTScore &  BERTScore\\
        &  en-en & en-de & en-it* & en-zh & en-en & en-en \\
    \midrule
    $\sim$no pretrain & 25.1\phantom{0}	& 72.49	& 73.61	& 76.93	& 80.88 & 83.89 \\
    $\sim$ASR pretrain & 21.42	& 76.72	& 79.73	& 80.62	& 82.48 & 85.97\\
    $\sim$contr. cos. &  \textbf{18.82}	& 77.31	& 80.27	& \textbf{80.76}	& 82.53 & 86.07\\
    $\sim$contr. wasser. & 19.07	& \textbf{77.33}	& 80.06	& 81.34	& \textbf{82.66} & \textbf{86.6}\phantom{0}\\
    
    \bottomrule
    \multicolumn{7}{c}{} \\
    \multicolumn{7}{l}{$\sim$ \phantom{H}  Model not trained on multilingual SSUM and SQA } \\
    \multicolumn{2}{l}{\gold \phantom{H}  Gold segmentation } & 
    \multicolumn{4}{l}{\nosplit \phantom{H} No segmentation (full audio used)}\\

    \end{tabular}%
    \caption{Ablation studies on different pretraining methods for the instruction following task: No pretraining, ASR pretraining and contrastive pretraining with either cosine similarity (\textit{contr. cos.}) or Wasserstein distance (\textit{contr. wasser.}). Test sets marked with * are automatically generated due to lack of availability for this language pair (see \cref{subsec:if_data}).}
    \label{tab:IF_ablation_contr}
\end{table*}
We explore three training strategies: (1) Direct fine-tuning on all available training data, (2)   ASR pretraining followed by fine-tuning, and (3) contrastive pretraining, as proposed by \citet{züfle2024contrastivelearningtaskindependentspeechllmpretraining}, followed by fine-tuning.

For contrastive pretraining, we use ASR data and experiment with cosine similarity and Wasserstein loss functions \citep{peyre-ot-2019, le2023pretrainingspeechtranslationctc}.  As shown in \cref{tab:IF_ablation_contr}, contrastive pretraining yields notable improvements over the other training strategies. Consequently, this approach is adopted for the final model submissions. Hyperparameter details are given in \cref{tab:IF_hyperparameter} in \cref{app:if_hyperparams}.

During initial experiments, our model struggled to distinguish answerable from unanswerable SQA questions. To improve this, we apply chain-of-thought prompting: the model first tags the question as answerable or not, then generates an answer only if applicable. This stepwise approach improves both classification and answer quality.

\subsection{Handling long audio}\label{subsec:if_long_audio}
\begin{table}
    \centering
    \resizebox{\linewidth}{!}{%
    \begin{tabular}{cccccc}
    \toprule
        \multirow{2}{*}{\textbf{Segm.}} & \multirow{2}{*}{\textbf{max secs.}} & \textbf{ASR} (WER) & \multicolumn{3}{c}{\textbf{ST} (COMET)}  \\
        & & en-en & en-de & en-it* & en-zh \\
        \midrule
        \gold & N/A &  18.77	& 77.15	& 80.65	& 81.83 \\
        \midrule
        \vad & 5 & 45.52 & 57.55	& 51.47	& 72.73\\
        \vad & 10 & 20.73	& 65.55	& 56.88	& 76.97\\        
        \vad & 15 & 20.74	& 68.92	& 58.24	& 77.44 \\
        \vad & 20 &\textbf{20.63}	& 69.94	& 59.01	& 77.45\\
        \vad & 25 & 25.48	& \textbf{71.61}	& \textbf{75.74}	& \textbf{78.04}\\
        \vad & 30 & -	& 70.79	& 58.99	& 76.16 \\
        \vad & 35 & -	& 67.54	& 56.88	& 76.5 \\
        \bottomrule
  \multicolumn{3}{l}{\gold \phantom{H}  Gold segmentation } &  \multicolumn{3}{l}{\vad \phantom{H} VAD segmentation} \\
    \end{tabular}%
    }
    \caption{Ablation study on Voice Activity Detection (VAD) segmentation using the \textit{IF contr. cos. model.} on the ACL 60/60 dataset. Test sets marked with * are automatically generated due to lack of availability for this language pair (see \cref{subsec:if_data}). For ASR, segmenting audio into chunks of up to 20 seconds yields the best results, while for ST, 25-second chunks perform best.}
    \label{tab:IF_ablation_vad}

\end{table}

The IF Constrained Long track involves processing audio inputs from five to ten minutes in duration.

\paragraph{ASR and ST} Initial experiments revealed that our model struggled with full-length audio inputs for ASR and ST, even when trained with artificially concatenated long-form sequences. To address this, we segment the input audio prior to inference. 

We use a Voice Activity Detection (VAD) approach \citep{vad1999} to segment audio, as due to track constraints, SHAS \citep{tsiamas22_interspeech} is not permitted.  For ASR, segmenting into chunks of up to 20 seconds yields best performance and for ST, segments of up to 25 seconds are more effective. Ablation results are provided in \cref{tab:IF_ablation_vad}.

\paragraph{SQA and SSUM}
For SQA and SSUM, we use the full audio. To handle long-form audio, we segment audio into 60-second chunks. Each chunk is  encoded, and the embeddings are concatenated before being passed to the Q-Former and LLM, following \citet{züfle2025nutshelldatasetabstractgeneration}. This strategy maintains full end-to-end trainability. For audios exceeding 26.7 minutes, we truncate the input to fit within memory constraints.

\subsection{Post-Editing}\label{subsec:if_postediting}
\begin{table}
    \centering
    \resizebox{\linewidth}{!}{%
    \begin{tabular}{ccccc}
    \toprule

        \multirow{2}{*}{\textbf{Post-editing context}} & \textbf{ASR} (WER) & \multicolumn{3}{c}{\textbf{ST} (COMET)}  \\
         & en-en & en-de & en-it* & en-zh \\
        \midrule
        No Post-Editing \vad & 20.63	& 71.61	& 75.74	& \textbf{78.04} \\
        \midrule
        1 \vad & 21.09	& 70.54	& 75.0\phantom{0}	& 77.22\\
        3 \vad &  20.96	& 71.91	& \textbf{75.88}	& 77.17\\        
        5 \vad & \textbf{20.43}	& 71.64	& 75.69& 	77.20 \\
        10 \vad & 21.88	& 71.90	& 75.53	& 77.14\\
        15 \vad & 50.07	& \textbf{71.95	}& \textbf{75.88} &	77.19 \\
        20 \vad & 50.12	& 71.82	& 75.55	& 77.20 \\

        \bottomrule
  \multicolumn{5}{l}{\vad \phantom{H}  VAD segmentation} \\
    \end{tabular}%
    }
    \caption{Ablation study on the context size of the postediting model using the \textit{IF contr. cos. model.} on the ACL 60/60 dataset. For ASR, a context size of 5 yields the best results, for ST, a context size of 15. For en$\to$zh, post-editing does not lead to an improvement.}
    \label{tab:IF_ablation_postediting}

\end{table}

\begin{table*}[t]
    \centering
    \resizebox{\linewidth}{!}{%
    \begin{tabular}{l|c|ccc|c|cccc|cccc}

    \toprule
     &  \textbf{ASR}\gold\vad & \multicolumn{3}{c|}{\textbf{ST}\gold\vad } & \textbf{SQA}\nosplit & \multicolumn{4}{c|}{\textbf{SQA}\nosplit} & \multicolumn{4}{c}{\textbf{SSUM}\nosplit} \\

       \textbf{Model}  &  ACL 60/60 & \multicolumn{3}{c|}{ACL 60/60} &  Sp.-SQuAD & \multicolumn{4}{c|}{NUTSHELL} & \multicolumn{4}{c}{NUTSHELL}\\
        &  WER & \multicolumn{3}{c|}{COMET} &  BERTScore & \multicolumn{4}{c|}{BERTScore} & \multicolumn{4}{c}{BERTScore}\\
        &  en-en & en-de & en-it* & en-zh & en-en & en-en* & en-de* & en-it* & en-zh* & en-en & en-de* & en-it* & en-zh* \\
    \midrule
        Phi-4\footref{foot:phi} &  16.8\phantom{0}\gold	& \textbf{79.19}\gold	& \textbf{83.43}\gold	& \textbf{83.23}\gold	&  82.56	& \textbf{91.78}	& \textbf{76.85}	& \textbf{78.41}	& \textbf{74.41}	& 86.27 & \textbf{67.71}	& \textbf{69.76}	 & \textbf{57.03} \\
        Qwen2 Audio\footref{foot:qwen}  &  20.14\gold	& 72.35\gold & 	74.23\gold	&77.19\gold	&  \textbf{87.35}	&89.0\phantom{0}	&71.81	&73.7&	69.62 &	84.88 &	63.06	& 64.17	& 51.79\\
        Whisper\footref{foot:whisper} + Llama 3.1\footref{foot:llama}   &  \textbf{14.67}\gold& 	78.04\gold	& 81.93\gold	& 77.66\gold	&  82.39	& 91.18	& 71.87 &	72.58	& 54.62 &	\textbf{86.62}	& 57.16	& 58.64 & 49.31 \\
        Seamless V2\footref{foot:seamless}  &  18.91\gold	& 73.64\gold & 	78.78\gold & 	75.26\gold &  -- & -- & -- & -- & -- & -- & -- & -- & -- \\
    \midrule
    IF contr. cos. &  18.77\gold	& \textbf{77.15}\gold	& 80.65\gold	& \textbf{81.83}\gold	&  82.83	& 93.04	& 79.73	& 82.08	& 79.8	& 86.83	& 68.46	& 71.01	&71.22\\
    IF contr. cos. tag &   19.82\gold	& 76.95\gold	& \textbf{80.69}\gold	& 81.75\gold	&  \textbf{82.86}	& 93.17	& 80.81	& 82.49	& \textbf{80.53}	& 86.52	& 68.31	& 71.06 &	71.09 \\
    IF contr. wasser. &   17.93\gold	 &72.47\gold	 &79.12\gold	& 80.88\gold	&  82.79	& \textbf{93.42}	& 80.65	& 82.46	& 80.47	& 86.86	& 68.45	& 71.08	& 71.37 \\
    IF contr. wasser. tag &  \textbf{17.78}\gold	& 74.06\gold	& 78.87\gold	& 81.10\gold & 	82.80	& 93.24	& \textbf{80.87}	& \textbf{82.76}  &	80.32	& \textbf{86.89}	& \textbf{68.76}	& \textbf{71.16}	& \textbf{71.54}\\
    \midrule

     IF contr. cos. &   20.63\vad	& 71.61\vad	& 75.74\vad	& 78.04\vad &  82.83 & 93.04	& 79.73	& 82.08	& 79.8	& 86.83	& 68.46	& 71.01	&71.22 \\
    \quad  + post-edit &  20.43\vad	& 71.95\vad	& 75.88\vad	& 77.19\vad	& \texttimes & \multicolumn{4}{c|}{\texttimes}& 86.85	& 68.61	& 71.22	& 71.17 \\

    IF contr. cos. tag &  33.24\vad	& 69.37\vad	& 73.36\vad	& 75.83\vad &  \textbf{82.86} & 93.17	& 80.81	& 82.49	& \textbf{80.53}	& 86.52	& 68.31	& 71.06 &	71.09  \\
    \quad  + post-edit &  33.53\vad	& 70.39\vad& 	73.14\vad	& 73.20\vad	&  \texttimes & \multicolumn{4}{c|}{\texttimes} & 86.54	& 68.42 &	71.16	 &71.01\\

    IF contr. wasser. &  21.88\vad	& 71.61\vad	& 76.78\vad & 	\textbf{78.21}\vad  &  82.79 & \textbf{93.42}	& 80.65	& 82.46	& 80.47	& 86.86	& 68.45	& 71.08	& 71.37 \\
    \quad  + post-edit &  33.51\vad	& 71.12\vad	& \textbf{77.23}\vad	& 68.02\vad	&  \texttimes &  \multicolumn{4}{c|}{\texttimes} & 86.88	& 68.68	& 71.12	& 71.24\\

    IF contr. wasser. tag  & 22.07\vad	& 71.84\vad	& 76.29\vad	& 78.24\vad  &   82.80	& 93.24	& \textbf{80.87}	& \textbf{82.76}  &	80.32	& 86.89	& 68.76	& 71.16	& \textbf{71.54}  \\
    \quad  + post-edit &  \textbf{19.76}\vad	& \textbf{72.29}\vad	& 76.75\vad	& 77.21\vad	&  \texttimes & \multicolumn{4}{c|}{\texttimes} &\textbf{ 86.90} & 	\textbf{68.95}	& \textbf{71.30}	& 71.41\\
    
    \bottomrule
    \multicolumn{14}{c}{} \\

    \multicolumn{3}{l}{\gold \phantom{H}  Gold segmentation } & \multicolumn{6}{l}{\vad \phantom{H}  Voice Activity Detection (VAD) segmentation} &  \multicolumn{4}{l}{\nosplit \phantom{H} No segmentation (full audio used)}\\
    \multicolumn{3}{l}{--\phantom{H}  Not supported by model}  & \multicolumn{6}{l}{\texttimes \phantom{H} post-editing not applied, because context is not available} \\

    \end{tabular}%
    }
    \caption{Results for baseline models and our end-to-end trained instruction-following models (\textit{IF}), developed for the Constraint Instruction Following Long track. The IF models are pretrained using contrastive learning, with either cosine similarity (\textit{contr. cos.}) or Wasserstein distance (\textit{contr. wasser.}). To improve performance on question answering, we also experiment with tagging answers to indicate whether the question is answerable (\textit{+ tag}). Test sets marked with * are automatically generated due to lack of availability for this language pair and task (see \cref{subsec:if_data}). The \textit{IF contr. wasser. tag + post-edit} model was submitted to the shared task. }
    \label{tab:if_long}

\end{table*}
To improve output quality, we use a post-editing model that works on document level. This helps to correct scientific terminology and it restores contextual coherence that may be lost due to segmentation of long audio inputs.

For ASR, we train the post-editing model on the SeamlessM4T-Large transcriptions of the TTS-generated scientific abstracts from NUTSHELL, paired with the original text.  For ST, we use the ACL 60/60 development set, transcribed by our IF model. The post-editing model setup is adapted from \cref{subsec:SHORT_asrpostedit}, with two key differences: in compliance with the constrained setting, we use LLaMA-3.1-8B-Instruct\footref{foot:llama} \citep{grattafiori2024llama3herdmodels} as the base model, and we predict the reference using only a single system output, since in the IF track we do not employ an ensemble.

We conduct experiments to examine the effect of context size on post-editing performance. For ASR, a context window of five sentences provides the best results, while ST benefits from a 15-sentence context. For en$\to$zh, no performance gains are achieved. These results are summarized in \cref{tab:IF_ablation_postediting}.  We also apply the post-editing model to SSUM outputs, using the full summary as context.

\subsection{Baselines}\label{subsec:IF_baselines}
We compare our system to four baseline models. We include two end-to-end Speech-LLMs:  Phi-4\footnote{\label{foot:phi}microsoft/Phi-4-multimodal-instruct} \citep{abdin2024phi4technicalreport} and Qwen2 Audio\footnote{\label{foot:qwen}Qwen/Qwen2-Audio-7B-Instruct} \citep{Qwen2-Audio}, using default parameter settings provided on Hugging Face model cards and following the prompts specified by the shared task. 
We also evaluate a cascaded baseline using Whisper-large-v3\footnote{\label{foot:whisper}openai/whisper-large-v3} \citep{radford2023robust}, and LLaMA-3.1-8B-Instruct\footref{foot:llama} \citep{grattafiori2024llama3herdmodels} to follow the instructions. 
Lastly, for ASR and ST, we include  SeamlessM4T-Large\footref{foot:seamless} \citep{seamlessm4t2023}, given that it also serves as the speech encoder in our own end-to-end architecture.

\subsection{Evaluation} We evaluate ASR with WER using JiWER, ST using  COMET\footnote{Unbabel/wmt22-comet-da} \citep{rei-etal-2022-comet}, and SQA and SSUM using BERTScore \citep{DBLP:conf/iclr/ZhangKWWA20}.  

\subsection{Results}\label{subsec:if_results}

All results can be found in \cref{tab:if_long}. We evaluate our approach against the baselines from  \cref{subsec:IF_baselines}, as well as four end-to-end trained instruction-following models (\textit{IF}). Among these, we compare two contrastive pretraining strategies (\textit{contr. cos.} and \textit{contr. wasser.}), as outlined in \cref{subsec:if_model}. For the SQA task, we also explore a chain-of-thought variant (\textit{tag}), as detailed in \cref{subsec:if_model}.

\paragraph{ASR and ST} 
Using gold segmentation, we compare our \textit{IF} models against the baselines. Phi-4\footref{foot:phi} \citep{abdin2024phi4technicalreport} achieves the strongest performance on ST, while Whisper\footref{foot:whisper} \citep{radford2023robust} performs best for ASR. However, our \textit{IF} models consistently outperform both Qwen2 Audio\footref{foot:qwen} \citep{Qwen2-Audio} and SeamlessM4T-Large\footref{foot:seamless} \citep{seamlessm4t2023}. The latter result confirms that our end-to-end architecture is able to improve over the speech foundation model. 

Under VAD segmentation, which is also used for the shared task testset, we observe a performance drop across all \textit{IF} models, as expected.
Applying post-editing partially mitigates this drop. For ASR, post-editing only improves \textit{IF contr. cos} and \textit{IF contr. wasser. tag}, bringing them close to their gold-segmented counterparts. In ST, post-editing yields consistent improvements for en$\to$de and en$\to$it, but not for en$\to$zh, likely due to the limited Chinese capabilities of the post-editing model and sparse training data in that language.

\paragraph{SQA and SSUM}
On the SQA-NUTSHELL dataset, all \textit{IF} models outperform the baselines, whereas on Spoken-SQuAD (which is extractive and out-of-domain), this is not the case. For SSUM, \textit{IF} models consistently surpass the baselines, particularly in en$\to$it and en$\to$zh. Post-editing yields slight gains for SSUM as well, though similar to ST, no improvement is observed for en$\to$zh.

\paragraph{Final Model} We select \textit{IF contr. wasser. tag + post-edit} for our final submission. It offers the best  performance for ASR, SQA, and SSUM, and is competitive with the other \textit{IF} models in ST.

\section{Conclusion}

This system paper presents KIT's submissions to the Offline and the IF Long tracks. By integrating LLMs into both cascaded and end-to-end architectures for speech processing, we demonstrate their potential in handling a range of spoken language tasks. For future work, we aim to explore a unified architecture capable of producing high-quality translations while also supporting instruction-following capabilities.

\section*{Acknowledgments}
Part of this work received support from the European Union’s Horizon research and innovation programme under grant agreement No 101135798, project Meetween (My Personal AI Mediator for Virtual MEETtings BetWEEN People). Part of this work was performed on the HoreKa supercomputer funded by the Ministry of Science, Research and the Arts Baden-Württemberg and by the Federal Ministry of Education and Research.

\bibliography{custom, anthology}

\begin{thebibliography}{48}
\providecommand{\natexlab}[1]{#1}

\bibitem[{Abdin et~al.(2024)Abdin, Aneja, Behl, Bubeck, Eldan, Gunasekar, Harrison, Hewett, Javaheripi, Kauffmann, Lee, Lee, Li, Liu, Mendes, Nguyen, Price, de~Rosa, Saarikivi, Salim, Shah, Wang, Ward, Wu, Yu, Zhang, and Zhang}]{abdin2024phi4technicalreport}
Marah Abdin, Jyoti Aneja, Harkirat Behl, Sébastien Bubeck, Ronen Eldan, Suriya Gunasekar, Michael Harrison, Russell~J. Hewett, Mojan Javaheripi, Piero Kauffmann, James~R. Lee, Yin~Tat Lee, Yuanzhi Li, Weishung Liu, Caio C.~T. Mendes, Anh Nguyen, Eric Price, Gustavo de~Rosa, Olli Saarikivi, and 8 others. 2024.
\newblock \href {https://arxiv.org/abs/2412.08905} {Phi-4 technical report}.
\newblock \emph{Preprint}, arXiv:2412.08905.

\bibitem[{Abouelenin et~al.(2025)Abouelenin, Ashfaq, Atkinson, Awadalla, Bach, Bao, Benhaim, Cai, Chaudhary, Chen et~al.}]{abouelenin2025phi}
Abdelrahman Abouelenin, Atabak Ashfaq, Adam Atkinson, Hany Awadalla, Nguyen Bach, Jianmin Bao, Alon Benhaim, Martin Cai, Vishrav Chaudhary, Congcong Chen, and 1 others. 2025.
\newblock Phi-4-mini technical report: Compact yet powerful multimodal language models via mixture-of-loras.
\newblock \emph{arXiv preprint arXiv:2503.01743}.

\bibitem[{Ahmad et~al.(2024)Ahmad, Anastasopoulos, Bojar, Borg, Carpuat, Cattoni, Cettolo, Chen, Dong, Federico, Haddow, Javorsk{\'y}, Krubi{\'n}ski, Lam, Ma, Mathur, Matusov, Maurya, McCrae, Murray, Nakamura, Negri, Niehues, Niu, Ojha, Ortega, Papi, Pol{\'a}k, Posp{\'i}{\v{s}}il, Pecina, Salesky, Sethiya, Sarkar, Shi, Sikasote, Sperber, St{\"u}ker, Sudoh, Thompson, Waibel, Watanabe, Wilken, Zem{\'a}nek, and Zevallos}]{ahmad-etal-2024-findings}
Ibrahim~Said Ahmad, Antonios Anastasopoulos, Ond{\v{r}}ej Bojar, Claudia Borg, Marine Carpuat, Roldano Cattoni, Mauro Cettolo, William Chen, Qianqian Dong, Marcello Federico, Barry Haddow, D{\'a}vid Javorsk{\'y}, Mateusz Krubi{\'n}ski, Tsz~Kin Lam, Xutai Ma, Prashant Mathur, Evgeny Matusov, Chandresh Maurya, John McCrae, and 25 others. 2024.
\newblock \href {https://doi.org/10.18653/v1/2024.iwslt-1.1} {{FINDINGS} {OF} {THE} {IWSLT} 2024 {EVALUATION} {CAMPAIGN}}.
\newblock In \emph{Proceedings of the 21st International Conference on Spoken Language Translation (IWSLT 2024)}, pages 1--11, Bangkok, Thailand (in-person and online). Association for Computational Linguistics.

\bibitem[{Alves et~al.(2024)Alves, Pombal, Guerreiro, Martins, Alves, Farajian, Peters, Rei, Fernandes, Agrawal et~al.}]{alvestower}
Duarte~Miguel Alves, Jos{\'e} Pombal, Nuno~M Guerreiro, Pedro~Henrique Martins, Jo{\~a}o Alves, Amin Farajian, Ben Peters, Ricardo Rei, Patrick Fernandes, Sweta Agrawal, and 1 others. 2024.
\newblock Tower: An open multilingual large language model for translation-related tasks.
\newblock In \emph{First Conference on Language Modeling}.

\bibitem[{Bain et~al.(2023)Bain, Huh, Han, and Zisserman}]{bain2022whisperx}
Max Bain, Jaesung Huh, Tengda Han, and Andrew Zisserman. 2023.
\newblock Whisperx: Time-accurate speech transcription of long-form audio.
\newblock \emph{INTERSPEECH 2023}.

\bibitem[{Bredin and Laurent(2021)}]{bredin21_interspeech}
Hervé Bredin and Antoine Laurent. 2021.
\newblock \href {https://doi.org/10.21437/Interspeech.2021-560} {End-to-end speaker segmentation for overlap-aware resegmentation}.
\newblock In \emph{Interspeech 2021}, pages 3111--3115.

\bibitem[{Chu et~al.(2024{\natexlab{a}})Chu, Xu, Yang, Wei, Wei, Guo, Leng, Lv, He, Lin, Zhou, and Zhou}]{Qwen2-Audio}
Yunfei Chu, Jin Xu, Qian Yang, Haojie Wei, Xipin Wei, Zhifang Guo, Yichong Leng, Yuanjun Lv, Jinzheng He, Junyang Lin, Chang Zhou, and Jingren Zhou. 2024{\natexlab{a}}.
\newblock Qwen2-audio technical report.
\newblock \emph{arXiv preprint arXiv:2407.10759}.

\bibitem[{Chu et~al.(2024{\natexlab{b}})Chu, Xu, Yang, Wei, Wei, Guo, Leng, Lv, He, Lin et~al.}]{chu2024qwen2}
Yunfei Chu, Jin Xu, Qian Yang, Haojie Wei, Xipin Wei, Zhifang Guo, Yichong Leng, Yuanjun Lv, Jinzheng He, Junyang Lin, and 1 others. 2024{\natexlab{b}}.
\newblock Qwen2-audio technical report.
\newblock \emph{arXiv preprint arXiv:2407.10759}.

\bibitem[{Communication et~al.(2023)Communication, Barrault, Chung, Meglioli, Dale, Dong, Duquenne, Elsahar, Gong, Heffernan, Hoffman, Klaiber, Li, Licht, Maillard, Rakotoarison, Sadagopan, Wenzek, Ye, Akula, Chen, Hachem, Ellis, Gonzalez, Haaheim, Hansanti, Howes, Huang, Hwang, Inaguma, Jain, Kalbassi, Kallet, Kulikov, Lam, Li, Ma, Mavlyutov, Peloquin, Ramadan, Ramakrishnan, Sun, Tran, Tran, Tufanov, Vogeti, Wood, Yang, Yu, Andrews, Balioglu, Costa-jussà, Celebi, Elbayad, Gao, Guzmán, Kao, Lee, Mourachko, Pino, Popuri, Ropers, Saleem, Schwenk, Tomasello, Wang, Wang, and Wang}]{seamlessm4t2023}
Seamless Communication, Loïc Barrault, Yu-An Chung, Mariano~Cora Meglioli, David Dale, Ning Dong, Paul-Ambroise Duquenne, Hady Elsahar, Hongyu Gong, Kevin Heffernan, John Hoffman, Christopher Klaiber, Pengwei Li, Daniel Licht, Jean Maillard, Alice Rakotoarison, Kaushik~Ram Sadagopan, Guillaume Wenzek, Ethan Ye, and 49 others. 2023.
\newblock \href {https://arxiv.org/abs/2308.11596} {Seamlessm4t: Massively multilingual \& multimodal machine translation}.
\newblock \emph{Preprint}, arXiv:2308.11596.

\bibitem[{Cui et~al.(2025)Cui, Gao, Liu, Luan, and Wang}]{cui2025multilingual}
Menglong Cui, Pengzhi Gao, Wei Liu, Jian Luan, and Bin Wang. 2025.
\newblock Multilingual machine translation with open large language models at practical scale: An empirical study.
\newblock \emph{arXiv preprint arXiv:2502.02481}.

\bibitem[{Finkelstein et~al.(2024)Finkelstein, Vilar, and Freitag}]{finkelstein-etal-2024-introducing}
Mara Finkelstein, David Vilar, and Markus Freitag. 2024.
\newblock \href {https://doi.org/10.18653/v1/2024.wmt-1.126} {Introducing the {N}ews{P}a{LM} {MBR} and {QE} dataset: {LLM}-generated high-quality parallel data outperforms traditional web-crawled data}.
\newblock In \emph{Proceedings of the Ninth Conference on Machine Translation}, pages 1355--1372, Miami, Florida, USA. Association for Computational Linguistics.

\bibitem[{Grattafiori et~al.(2024)Grattafiori, Dubey, Jauhri, Pandey, Kadian, Al-Dahle, Letman, Mathur, Schelten, Vaughan, Yang, Fan, Goyal, Hartshorn, Yang, Mitra, Sravankumar, Korenev, Hinsvark, Rao, Zhang, Rodriguez, Gregerson, Spataru, Roziere, Biron, Tang, Chern, Caucheteux, Nayak, Bi, Marra, McConnell, Keller, Touret, Wu, Wong, Ferrer, Nikolaidis, Allonsius, Song, Pintz, Livshits, Wyatt, Esiobu, Choudhary, Mahajan, Garcia-Olano, Perino, Hupkes, Lakomkin, AlBadawy, Lobanova, Dinan, Smith, Radenovic, Guzmán, Zhang, Synnaeve, Lee, Anderson, Thattai, Nail, Mialon, Pang, Cucurell, Nguyen, Korevaar, Xu, Touvron, Zarov, Ibarra, Kloumann, Misra, Evtimov, Zhang, Copet, Lee, Geffert, Vranes, Park, Mahadeokar, Shah, van~der Linde, Billock, Hong, Lee, Fu, Chi, Huang, Liu, Wang, Yu, Bitton, Spisak, Park, Rocca, Johnstun, Saxe, Jia, Alwala, Prasad, Upasani, Plawiak, Li, Heafield, Stone, El-Arini, Iyer, Malik, Chiu, Bhalla, Lakhotia, Rantala-Yeary, van~der Maaten, Chen, Tan, Jenkins, Martin, Madaan, Malo, Blecher,
  Landzaat, de~Oliveira, Muzzi, Pasupuleti, Singh, Paluri, Kardas, Tsimpoukelli, Oldham, Rita, Pavlova, Kambadur, Lewis, Si, Singh, Hassan, Goyal, Torabi, Bashlykov, Bogoychev, Chatterji, Zhang, Duchenne, Çelebi, Alrassy, Zhang, Li, Vasic, Weng, Bhargava, Dubal, Krishnan, Koura, Xu, He, Dong, Srinivasan, Ganapathy, Calderer, Cabral, Stojnic, Raileanu, Maheswari, Girdhar, Patel, Sauvestre, Polidoro, Sumbaly, Taylor, Silva, Hou, Wang, Hosseini, Chennabasappa, Singh, Bell, Kim, Edunov, Nie, Narang, Raparthy, Shen, Wan, Bhosale, Zhang, Vandenhende, Batra, Whitman, Sootla, Collot, Gururangan, Borodinsky, Herman, Fowler, Sheasha, Georgiou, Scialom, Speckbacher, Mihaylov, Xiao, Karn, Goswami, Gupta, Ramanathan, Kerkez, Gonguet, Do, Vogeti, Albiero, Petrovic, Chu, Xiong, Fu, Meers, Martinet, Wang, Wang, Tan, Xia, Xie, Jia, Wang, Goldschlag, Gaur, Babaei, Wen, Song, Zhang, Li, Mao, Coudert, Yan, Chen, Papakipos, Singh, Srivastava, Jain, Kelsey, Shajnfeld, Gangidi, Victoria, Goldstand, Menon, Sharma, Boesenberg,
  Baevski, Feinstein, Kallet, Sangani, Teo, Yunus, Lupu, Alvarado, Caples, Gu, Ho, Poulton, Ryan, Ramchandani, Dong, Franco, Goyal, Saraf, Chowdhury, Gabriel, Bharambe, Eisenman, Yazdan, James, Maurer, Leonhardi, Huang, Loyd, Paola, Paranjape, Liu, Wu, Ni, Hancock, Wasti, Spence, Stojkovic, Gamido, Montalvo, Parker, Burton, Mejia, Liu, Wang, Kim, Zhou, Hu, Chu, Cai, Tindal, Feichtenhofer, Gao, Civin, Beaty, Kreymer, Li, Adkins, Xu, Testuggine, David, Parikh, Liskovich, Foss, Wang, Le, Holland, Dowling, Jamil, Montgomery, Presani, Hahn, Wood, Le, Brinkman, Arcaute, Dunbar, Smothers, Sun, Kreuk, Tian, Kokkinos, Ozgenel, Caggioni, Kanayet, Seide, Florez, Schwarz, Badeer, Swee, Halpern, Herman, Sizov, Guangyi, Zhang, Lakshminarayanan, Inan, Shojanazeri, Zou, Wang, Zha, Habeeb, Rudolph, Suk, Aspegren, Goldman, Zhan, Damlaj, Molybog, Tufanov, Leontiadis, Veliche, Gat, Weissman, Geboski, Kohli, Lam, Asher, Gaya, Marcus, Tang, Chan, Zhen, Reizenstein, Teboul, Zhong, Jin, Yang, Cummings, Carvill, Shepard, McPhie,
  Torres, Ginsburg, Wang, Wu, U, Saxena, Khandelwal, Zand, Matosich, Veeraraghavan, Michelena, Li, Jagadeesh, Huang, Chawla, Huang, Chen, Garg, A, Silva, Bell, Zhang, Guo, Yu, Moshkovich, Wehrstedt, Khabsa, Avalani, Bhatt, Mankus, Hasson, Lennie, Reso, Groshev, Naumov, Lathi, Keneally, Liu, Seltzer, Valko, Restrepo, Patel, Vyatskov, Samvelyan, Clark, Macey, Wang, Hermoso, Metanat, Rastegari, Bansal, Santhanam, Parks, White, Bawa, Singhal, Egebo, Usunier, Mehta, Laptev, Dong, Cheng, Chernoguz, Hart, Salpekar, Kalinli, Kent, Parekh, Saab, Balaji, Rittner, Bontrager, Roux, Dollar, Zvyagina, Ratanchandani, Yuvraj, Liang, Alao, Rodriguez, Ayub, Murthy, Nayani, Mitra, Parthasarathy, Li, Hogan, Battey, Wang, Howes, Rinott, Mehta, Siby, Bondu, Datta, Chugh, Hunt, Dhillon, Sidorov, Pan, Mahajan, Verma, Yamamoto, Ramaswamy, Lindsay, Lindsay, Feng, Lin, Zha, Patil, Shankar, Zhang, Zhang, Wang, Agarwal, Sajuyigbe, Chintala, Max, Chen, Kehoe, Satterfield, Govindaprasad, Gupta, Deng, Cho, Virk, Subramanian, Choudhury,
  Goldman, Remez, Glaser, Best, Koehler, Robinson, Li, Zhang, Matthews, Chou, Shaked, Vontimitta, Ajayi, Montanez, Mohan, Kumar, Mangla, Ionescu, Poenaru, Mihailescu, Ivanov, Li, Wang, Jiang, Bouaziz, Constable, Tang, Wu, Wang, Wu, Gao, Kleinman, Chen, Hu, Jia, Qi, Li, Zhang, Zhang, Adi, Nam, Yu, Wang, Zhao, Hao, Qian, Li, He, Rait, DeVito, Rosnbrick, Wen, Yang, Zhao, and Ma}]{grattafiori2024llama3herdmodels}
Aaron Grattafiori, Abhimanyu Dubey, Abhinav Jauhri, Abhinav Pandey, Abhishek Kadian, Ahmad Al-Dahle, Aiesha Letman, Akhil Mathur, Alan Schelten, Alex Vaughan, Amy Yang, Angela Fan, Anirudh Goyal, Anthony Hartshorn, Aobo Yang, Archi Mitra, Archie Sravankumar, Artem Korenev, Arthur Hinsvark, and 542 others. 2024.
\newblock \href {https://arxiv.org/abs/2407.21783} {The llama 3 herd of models}.
\newblock \emph{Preprint}, arXiv:2407.21783.

\bibitem[{Guerreiro et~al.(2024)Guerreiro, Rei, Stigt, Coheur, Colombo, and Martins}]{guerreiro2024xcomet}
Nuno~M Guerreiro, Ricardo Rei, Daan~van Stigt, Luisa Coheur, Pierre Colombo, and Andr{\'e}~FT Martins. 2024.
\newblock xcomet: Transparent machine translation evaluation through fine-grained error detection.
\newblock \emph{Transactions of the Association for Computational Linguistics}, 12:979--995.

\bibitem[{Hu et~al.(2022)Hu, Wallis, Allen-Zhu, Li, Wang, Wang, Chen et~al.}]{hulora}
Edward~J Hu, Phillip Wallis, Zeyuan Allen-Zhu, Yuanzhi Li, Shean Wang, Lu~Wang, Weizhu Chen, and 1 others. 2022.
\newblock Lora: Low-rank adaptation of large language models.
\newblock In \emph{International Conference on Learning Representations}.

\bibitem[{{Iranzo-Sánchez} et~al.(2020){Iranzo-Sánchez}, {Silvestre-Cerdà}, {Jorge}, {Roselló}, {Giménez}, {Sanchis}, {Civera}, and {Juan}}]{jairsan2020a}
J.~{Iranzo-Sánchez}, J.~A. {Silvestre-Cerdà}, J.~{Jorge}, N.~{Roselló}, A.~{Giménez}, A.~{Sanchis}, J.~{Civera}, and A.~{Juan}. 2020.
\newblock Europarl-st: A multilingual corpus for speech translation of parliamentary debates.
\newblock In \emph{ICASSP 2020 - 2020 IEEE International Conference on Acoustics, Speech and Signal Processing (ICASSP)}, pages 8229--8233.

\bibitem[{Juraska et~al.(2024)Juraska, Deutsch, Finkelstein, and Freitag}]{juraska-etal-2024-metricx}
Juraj Juraska, Daniel Deutsch, Mara Finkelstein, and Markus Freitag. 2024.
\newblock \href {https://doi.org/10.18653/v1/2024.wmt-1.35} {{M}etric{X}-24: The {G}oogle submission to the {WMT} 2024 metrics shared task}.
\newblock In \emph{Proceedings of the Ninth Conference on Machine Translation}, pages 492--504, Miami, Florida, USA. Association for Computational Linguistics.

\bibitem[{Kim et~al.(2021)Kim, Kong, and Son}]{kim2021conditional}
Jaehyeon Kim, Jungil Kong, and Juhee Son. 2021.
\newblock Conditional variational autoencoder with adversarial learning for end-to-end text-to-speech.
\newblock In \emph{International Conference on Machine Learning}, pages 5530--5540. PMLR.

\bibitem[{Koehn(2005)}]{koehn-2005-europarl}
Philipp Koehn. 2005.
\newblock \href {https://aclanthology.org/2005.mtsummit-papers.11/} {{E}uroparl: A parallel corpus for statistical machine translation}.
\newblock In \emph{Proceedings of Machine Translation Summit X: Papers}, pages 79--86, Phuket, Thailand.

\bibitem[{Koneru et~al.(2024{\natexlab{a}})Koneru, Binh~Nguyen, Pham, Liu, Li, Waibel, and Niehues}]{koneru-etal-2024-blending}
Sai Koneru, Thai Binh~Nguyen, Ngoc-Quan Pham, Danni Liu, Zhaolin Li, Alexander Waibel, and Jan Niehues. 2024{\natexlab{a}}.
\newblock \href {https://doi.org/10.18653/v1/2024.iwslt-1.24} {Blending {LLM}s into cascaded speech translation: {KIT}`s offline speech translation system for {IWSLT} 2024}.
\newblock In \emph{Proceedings of the 21st International Conference on Spoken Language Translation (IWSLT 2024)}, pages 183--191, Bangkok, Thailand (in-person and online). Association for Computational Linguistics.

\bibitem[{Koneru et~al.(2024{\natexlab{b}})Koneru, Exel, Huck, and Niehues}]{koneru-etal-2024-contextual}
Sai Koneru, Miriam Exel, Matthias Huck, and Jan Niehues. 2024{\natexlab{b}}.
\newblock \href {https://doi.org/10.18653/v1/2024.naacl-long.148} {Contextual refinement of translations: Large language models for sentence and document-level post-editing}.
\newblock In \emph{Proceedings of the 2024 Conference of the North American Chapter of the Association for Computational Linguistics: Human Language Technologies (Volume 1: Long Papers)}, pages 2711--2725, Mexico City, Mexico. Association for Computational Linguistics.

\bibitem[{Koneru et~al.(2025)Koneru, Huck, Exel, and Niehues}]{koneru2025quality}
Sai Koneru, Matthias Huck, Miriam Exel, and Jan Niehues. 2025.
\newblock Quality-aware decoding: Unifying quality estimation and decoding.
\newblock \emph{arXiv preprint arXiv:2502.08561}.

\bibitem[{Le et~al.(2023)Le, Gong, Wang, Pino, Lecouteux, and Schwab}]{le2023pretrainingspeechtranslationctc}
Phuong-Hang Le, Hongyu Gong, Changhan Wang, Juan Pino, Benjamin Lecouteux, and Didier Schwab. 2023.
\newblock \href {https://arxiv.org/abs/2301.11716} {Pre-training for speech translation: Ctc meets optimal transport}.
\newblock \emph{Preprint}, arXiv:2301.11716.

\bibitem[{Lee et~al.(2018)Lee, Wu, Liu, and Lee}]{lee2018spoken}
Chia-Hsuan Lee, Szu-Lin Wu, Chi-Liang Liu, and Hung-yi Lee. 2018.
\newblock Spoken squad: A study of mitigating the impact of speech recognition errors on listening comprehension.
\newblock \emph{Proc. Interspeech 2018}, pages 3459--3463.

\bibitem[{Lerner et~al.(2022)Lerner, Bergo{\"e}nd, Guinaudeau, Bredin, Maurice, Lefevre, Bouteiller, Berhe, Galmant, Yin, and Barras}]{lerner-etal-2022-bazinga}
Paul Lerner, Juliette Bergo{\"e}nd, Camille Guinaudeau, Herv{\'e} Bredin, Benjamin Maurice, Sharleyne Lefevre, Martin Bouteiller, Aman Berhe, L{\'e}o Galmant, Ruiqing Yin, and Claude Barras. 2022.
\newblock \href {https://aclanthology.org/2022.lrec-1.367/} {Bazinga! a dataset for multi-party dialogues structuring}.
\newblock In \emph{Proceedings of the Thirteenth Language Resources and Evaluation Conference}, pages 3434--3441, Marseille, France. European Language Resources Association.

\bibitem[{Li et~al.(2023)Li, Li, Savarese, and Hoi}]{Li2023BLIP2BL}
Junnan Li, Dongxu Li, Silvio Savarese, and Steven C.~H. Hoi. 2023.
\newblock \href {https://api.semanticscholar.org/CorpusID:256390509} {Blip-2: Bootstrapping language-image pre-training with frozen image encoders and large language models}.
\newblock In \emph{International Conference on Machine Learning}.

\bibitem[{Lison and Tiedemann(2016)}]{lison-tiedemann-2016-opensubtitles2016}
Pierre Lison and J{\"o}rg Tiedemann. 2016.
\newblock \href {https://aclanthology.org/L16-1147/} {{O}pen{S}ubtitles2016: Extracting large parallel corpora from movie and {TV} subtitles}.
\newblock In \emph{Proceedings of the Tenth International Conference on Language Resources and Evaluation ({LREC}`16)}, pages 923--929, Portoro{\v{z}}, Slovenia. European Language Resources Association (ELRA).

\bibitem[{Liu et~al.(2023)Liu, Binh~Nguyen, Koneru, Yavuz~Ugan, Pham, Nam~Nguyen, Anh~Dinh, Mullov, Waibel, and Niehues}]{liu-etal-2023-kits}
Danni Liu, Thai Binh~Nguyen, Sai Koneru, Enes Yavuz~Ugan, Ngoc-Quan Pham, Tuan Nam~Nguyen, Tu~Anh~Dinh, Carlos Mullov, Alexander Waibel, and Jan Niehues. 2023.
\newblock \href {https://doi.org/10.18653/v1/2023.iwslt-1.6} {{KIT}`s multilingual speech translation system for {IWSLT} 2023}.
\newblock In \emph{Proceedings of the 20th International Conference on Spoken Language Translation (IWSLT 2023)}, pages 113--122, Toronto, Canada (in-person and online). Association for Computational Linguistics.

\bibitem[{Peyré and Cuturi(2019)}]{peyre-ot-2019}
Gabriel Peyré and Marco Cuturi. 2019.
\newblock \href {https://doi.org/10.1561/2200000073} {Computational optimal transport: With applications to data science}.
\newblock \emph{Foundations and Trends® in Machine Learning}, 11:355--206.

\bibitem[{Popovi{\'c}(2015)}]{popovic-2015-chrf}
Maja Popovi{\'c}. 2015.
\newblock \href {https://doi.org/10.18653/v1/W15-3049} {chr{F}: character n-gram {F}-score for automatic {MT} evaluation}.
\newblock In \emph{Proceedings of the Tenth Workshop on Statistical Machine Translation}, pages 392--395, Lisbon, Portugal. Association for Computational Linguistics.

\bibitem[{Radford et~al.(2023)Radford, Kim, Xu, Brockman, McLeavey, and Sutskever}]{radford2023robust}
Alec Radford, Jong~Wook Kim, Tao Xu, Greg Brockman, Christine McLeavey, and Ilya Sutskever. 2023.
\newblock Robust speech recognition via large-scale weak supervision.
\newblock In \emph{International conference on machine learning}, pages 28492--28518. PMLR.

\bibitem[{Ramos et~al.(2024)Ramos, Fernandes, Farinhas, and Martins}]{ramos-etal-2024-aligning}
Miguel Ramos, Patrick Fernandes, Ant{\'o}nio Farinhas, and Andre Martins. 2024.
\newblock \href {https://aclanthology.org/2024.eamt-1.22/} {Aligning neural machine translation models: Human feedback in training and inference}.
\newblock In \emph{Proceedings of the 25th Annual Conference of the European Association for Machine Translation (Volume 1)}, pages 258--274, Sheffield, UK. European Association for Machine Translation (EAMT).

\bibitem[{Rei et~al.(2022{\natexlab{a}})Rei, C.~de Souza, Alves, Zerva, Farinha, Glushkova, Lavie, Coheur, and Martins}]{rei-etal-2022-comet}
Ricardo Rei, Jos{\'e}~G. C.~de Souza, Duarte Alves, Chrysoula Zerva, Ana~C Farinha, Taisiya Glushkova, Alon Lavie, Luisa Coheur, and Andr{\'e} F.~T. Martins. 2022{\natexlab{a}}.
\newblock \href {https://aclanthology.org/2022.wmt-1.52/} {{COMET}-22: Unbabel-{IST} 2022 submission for the metrics shared task}.
\newblock In \emph{Proceedings of the Seventh Conference on Machine Translation (WMT)}, pages 578--585, Abu Dhabi, United Arab Emirates (Hybrid). Association for Computational Linguistics.

\bibitem[{Rei et~al.(2022{\natexlab{b}})Rei, Treviso, Guerreiro, Zerva, Farinha, Maroti, C.~de Souza, Glushkova, Alves, Coheur, Lavie, and Martins}]{rei-etal-2022-cometkiwi}
Ricardo Rei, Marcos Treviso, Nuno~M. Guerreiro, Chrysoula Zerva, Ana~C Farinha, Christine Maroti, Jos{\'e}~G. C.~de Souza, Taisiya Glushkova, Duarte Alves, Luisa Coheur, Alon Lavie, and Andr{\'e} F.~T. Martins. 2022{\natexlab{b}}.
\newblock \href {https://aclanthology.org/2022.wmt-1.60/} {{C}omet{K}iwi: {IST}-unbabel 2022 submission for the quality estimation shared task}.
\newblock In \emph{Proceedings of the Seventh Conference on Machine Translation (WMT)}, pages 634--645, Abu Dhabi, United Arab Emirates (Hybrid). Association for Computational Linguistics.

\bibitem[{Salesky et~al.(2023)Salesky, Darwish, Al-Badrashiny, Diab, and Niehues}]{salesky-etal-2023-evaluating}
Elizabeth Salesky, Kareem Darwish, Mohamed Al-Badrashiny, Mona Diab, and Jan Niehues. 2023.
\newblock \href {https://doi.org/10.18653/v1/2023.iwslt-1.2} {Evaluating multilingual speech translation under realistic conditions with resegmentation and terminology}.
\newblock In \emph{Proceedings of the 20th International Conference on Spoken Language Translation (IWSLT 2023)}, pages 62--78, Toronto, Canada (in-person and online). Association for Computational Linguistics.

\bibitem[{Sohn et~al.(1999)Sohn, Kim, and Sung}]{vad1999}
Jongseo Sohn, Nam~Soo Kim, and Wonyong Sung. 1999.
\newblock \href {https://doi.org/10.1109/97.736233} {A statistical model-based voice activity detection}.
\newblock \emph{IEEE Signal Processing Letters}, 6(1):1--3.

\bibitem[{Tang et~al.(2024)Tang, Yu, Sun, Chen, Tan, Li, Lu, MA, and Zhang}]{tang2024salmonn}
Changli Tang, Wenyi Yu, Guangzhi Sun, Xianzhao Chen, Tian Tan, Wei Li, Lu~Lu, Zejun MA, and Chao Zhang. 2024.
\newblock \href {https://openreview.net/forum?id=14rn7HpKVk} {{SALMONN}: Towards generic hearing abilities for large language models}.
\newblock In \emph{The Twelfth International Conference on Learning Representations}.

\bibitem[{Tang et~al.(2023)Tang, Yu, Sun, Chen, Tan, Li, Lu, Zejun, and Zhang}]{tangsalmonn}
Changli Tang, Wenyi Yu, Guangzhi Sun, Xianzhao Chen, Tian Tan, Wei Li, Lu~Lu, MA~Zejun, and Chao Zhang. 2023.
\newblock Salmonn: Towards generic hearing abilities for large language models.
\newblock In \emph{The Twelfth International Conference on Learning Representations}.

\bibitem[{Team(2021)}]{SileroModels}
Silero Team. 2021.
\newblock Silero models: pre-trained enterprise-grade stt / tts models and benchmarks.
\newblock \url{https://github.com/snakers4/silero-models}.

\bibitem[{Tsiamas et~al.(2022)Tsiamas, Gállego, Fonollosa, and Costa-jussà}]{tsiamas22_interspeech}
Ioannis Tsiamas, Gerard~I. Gállego, José A.~R. Fonollosa, and Marta~R. Costa-jussà. 2022.
\newblock \href {https://doi.org/10.21437/Interspeech.2022-59} {{SHAS: Approaching optimal Segmentation for End-to-End Speech Translation}}.
\newblock In \emph{Proc. Interspeech 2022}, pages 106--110.

\bibitem[{Wang et~al.(2020)Wang, Wu, and Pino}]{wang2020covost}
Changhan Wang, Anne Wu, and Juan Pino. 2020.
\newblock \href {https://arxiv.org/abs/2007.10310} {Covost 2: A massively multilingual speech-to-text translation corpus}.
\newblock \emph{Preprint}, arXiv:2007.10310.

\bibitem[{Wiseman(2019)}]{wiseman2019wiseman}
John Wiseman. 2019.
\newblock Wiseman/py-webrtcvad.
\newblock \emph{GitHub repository, Nov}.

\bibitem[{Xu et~al.(2024{\natexlab{a}})Xu, Murray, Koehn, Hoang, Eriguchi, and Khayrallah}]{xu2024x}
Haoran Xu, Kenton Murray, Philipp Koehn, Hieu Hoang, Akiko Eriguchi, and Huda Khayrallah. 2024{\natexlab{a}}.
\newblock X-alma: Plug \& play modules and adaptive rejection for quality translation at scale.
\newblock \emph{arXiv preprint arXiv:2410.03115}.

\bibitem[{Xu et~al.(2024{\natexlab{b}})Xu, Sharaf, Chen, Tan, Shen, Van~Durme, Murray, and Kim}]{xu2024contrastive}
Haoran Xu, Amr Sharaf, Yunmo Chen, Weiting Tan, Lingfeng Shen, Benjamin Van~Durme, Kenton Murray, and Young~Jin Kim. 2024{\natexlab{b}}.
\newblock Contrastive preference optimization: Pushing the boundaries of llm performance in machine translation.
\newblock In \emph{International Conference on Machine Learning}, pages 55204--55224. PMLR.

\bibitem[{Yan et~al.(2024)Yan, Fernandes, Tian, Ouyang, Chen, Livescu, Li, Neubig, and Watanabe}]{yan-etal-2024-cmus}
Brian Yan, Patrick Fernandes, Jinchuan Tian, Siqi Ouyang, William Chen, Karen Livescu, Lei Li, Graham Neubig, and Shinji Watanabe. 2024.
\newblock \href {https://doi.org/10.18653/v1/2024.iwslt-1.22} {{CMU}`s {IWSLT} 2024 offline speech translation system: A cascaded approach for long-form robustness}.
\newblock In \emph{Proceedings of the 21st International Conference on Spoken Language Translation (IWSLT 2024)}, pages 164--169, Bangkok, Thailand (in-person and online). Association for Computational Linguistics.

\bibitem[{Zhang et~al.(2020)Zhang, Kishore, Wu, Weinberger, and Artzi}]{DBLP:conf/iclr/ZhangKWWA20}
Tianyi Zhang, Varsha Kishore, Felix Wu, Kilian~Q. Weinberger, and Yoav Artzi. 2020.
\newblock \href {https://openreview.net/forum?id=SkeHuCVFDr} {Bertscore: Evaluating text generation with {BERT}}.
\newblock In \emph{8th International Conference on Learning Representations, {ICLR} 2020, Addis Ababa, Ethiopia, April 26-30, 2020}. OpenReview.net.

\bibitem[{Z{\"u}fle et~al.(2025)Z{\"u}fle, Papi, Savoldi, Gaido, Bentivogli, and Niehues}]{zufle2025nutshell}
Maike Z{\"u}fle, Sara Papi, Beatrice Savoldi, Marco Gaido, Luisa Bentivogli, and Jan Niehues. 2025.
\newblock Nutshell: A dataset for abstract generation from scientific talks.
\newblock \emph{arXiv preprint arXiv:2502.16942}.

\bibitem[{Züfle and Niehues(2024)}]{züfle2024contrastivelearningtaskindependentspeechllmpretraining}
Maike Züfle and Jan Niehues. 2024.
\newblock \href {https://arxiv.org/abs/2412.15712} {Contrastive learning for task-independent speechllm-pretraining}.
\newblock \emph{Preprint}, arXiv:2412.15712.

\bibitem[{Züfle et~al.(2025)Züfle, Papi, Savoldi, Gaido, Bentivogli, and Niehues}]{züfle2025nutshelldatasetabstractgeneration}
Maike Züfle, Sara Papi, Beatrice Savoldi, Marco Gaido, Luisa Bentivogli, and Jan Niehues. 2025.
\newblock \href {https://arxiv.org/abs/2502.16942} {Nutshell: A dataset for abstract generation from scientific talks}.
\newblock \emph{Preprint}, arXiv:2502.16942.

\end{thebibliography}

\appendix
\section{Offline Track - Prompts}
\label{sec:fuseasrprompt}

\noindent \textbf{LLM Fuse Prompt}
\begin{lstlisting}[breaklines=true, breakindent=0pt]
Post-Edit the Automatic Speech Recognition Transcripts from different systems understanding the context.

ASR Transcripts:

System1: {Whisper v2 Hyps}
System2: {Whisper v2 FT Hyps}
System3: {Phi-4 Hyps}
System4: {Whisper v3 Hyps}

Post-Edited Transcript: 
{Reference}
\end{lstlisting}

\noindent \textbf{MT APE Prompt}
\begin{lstlisting}[breaklines=true, breakindent=0pt]
<|im_start|>user
Post-Edit the German Translation of the English sentence.
English:
{src}
German:
{mt}
<|im_end|>
<|im_start|>assistant
Post-Edited German:
{ref}
\end{lstlisting}

\section{Instruction-Following Track - Prompts}
\label{sec:if_appendix}
\subsection{Data Augmentation Prompts SQA}\label{app:data_augm_prompts}

\noindent System Prompt:
\begin{lstlisting}[breaklines=true, breakindent=0pt]
You are a professional question generator. Given a transcript, you will create three questions: 
two that can be answered based on the transcript and one that cannot be answered (but is relevant to the topic). 
The answers should be full sentences in the target language specified. 
Your response must be in valid JSON format, with keys for 'questions' and 'answers'. 
Do not include any explanations or additional text.\n
\end{lstlisting}
Prompt:
\begin{lstlisting}[breaklines=true, breakindent=0pt]
<Transcript>\n
Based on the transcript, generate a JSON dictionary with the following structure.
The questions and answers must be in <trg lang>:\n
{{\n
'  "questions": [\n'
'    {"q1": "First question in <trg lang>", "a1": "Full-sentence answer in <trg lang>"},\n'
'    {"q2": "Second question in <trg lang>", "a2": "Full-sentence answer in <trg lang>"},\n'
f'    {{"q3": "Third question in <trg lang>", "a3": "N/A"}}\n'
  ]\n
}}\n
Ensure the response is a valid JSON object with properly formatted keys and values.
\end{lstlisting}

\subsection{Data Augmentation Prompts SSUM}\label{app:data_augm_prompts_ssum}

\noindent System Prompt:
\begin{lstlisting}[breaklines=true, breakindent=0pt]
A chat between a curious user and a professional system for translating ACL abstracts.\n
\end{lstlisting}
Prompt:
\begin{lstlisting}[breaklines=true, breakindent=0pt]
<abstract>\nTranslate this abstract to <trg lang>. Do not provide any explanation or additional text.
\end{lstlisting}

\subsection{Data Augmentation Prompts ST}\label{app:data_augm_prompts_st}

\noindent System Prompt:
\begin{lstlisting}[breaklines=true, breakindent=0pt]
You are a professional translator. Your task is to provide accurate, fluent, and natural translations without adding explanations, comments, or extra content.
\end{lstlisting}
Prompt:
\begin{lstlisting}[breaklines=true, breakindent=0pt]
Translate the following English text into <trg lang>. Do not provide any explanation or additional text.\n<text>
\end{lstlisting}

\subsection{Hyperparameters Model Training}\label{app:if_hyperparams}
\begin{table}[ht]
    \centering
    \resizebox{\linewidth}{!}{%
    \begin{tabular}{llc}
    \toprule
        training & Q-Former Num Query Token & 4 \\
        parameters & Q-Former Num Hidden Layers & 4\\
         & Q-Former Num Attention Heads & 12 \\
         & Q-Former Seconds per Window & 1/3 \\
         &  num GPUs & 4 \\
         & learning rate & 1e-4 \\
         & warmup ratio & 0.03 \\
         & optimizer & adamw\_torch \\
         & learning rate scheduler type & cosine \\
         & model max length & 2048 \\
         & gradient clipping & 1 \\
    \midrule
         pretraining  & num epochs & 5 \\
         specific   & per device batch size & 10 \\
         & gradient accumulation steps & 2 \\
            & contrastive $\tau$ cos + wasser & 0.1 \\
         & contrastive $\tau$ nwp & 0.5 \\
         & sinkhorn loss p & 2 \\
         & sinkhorn loss blur & 0.5 \\
    \midrule
        finetuning  &  num epochs    & 2 \\
        specific & per device batch size & 2 \\
         & gradient accumulation steps & 10 \\
    \bottomrule
    \end{tabular}%
    }
    \caption{Hyperparameters for the trainings, which are conducted on four NVIDIA GH200 96GB GPUs, mostly following \citet{züfle2024contrastivelearningtaskindependentspeechllmpretraining}.}
    \label{tab:IF_hyperparameter}
\end{table}

\end{document}